# Potentially Guided Bidirectionalized RRT* for Fast Optimal Path Planning in Cluttered Environments

Zaid Tahir[1*], Ahmed H. Qureshi[1,2*], Yasar Ayaz[1*], Raheel Nawaz[3*]

[1] *Robotics And Intelligent Systems Engineering (RISE) Lab, Department of Robotics and Artificial Intelligence, School of Mechanical And Manufacturing Engineering (SMME), National University of Sciences And Technology (NUST), H-12 Campus, Islamabad, 44000, Pakistan.*

[2] *Department of Electrical and Computer Engineering, University of California San Diego, 9500 Gilman Dr, La Jolla, CA 92093.*

[3] *School of Computing, Mathematics and Digital Technology, Manchester Metropolitan University (MMU), Manchester M15 6BH, England.*

**Abstract**

Rapidly-exploring Random Tree star (RRT*) has recently gained immense popularity in the motion planning community as it provides a probabilistically complete and asymptotically optimal solution without requiring the complete information of the obstacle space. In spite of all of its advantages, RRT* converges to optimal solution very slowly. Hence to improve the convergence rate, its bidirectional variants were introduced, the Bi-directional RRT* (B-RRT*) and Intelligent Bi-directional RRT* (IB-RRT*). However, as both variants perform pure exploration, they tend to suffer in highly cluttered environments. In order to overcome these limitations we introduce a new concept of potentially guided bidirectional trees in our proposed Potentially Guided Intelligent Bi-directional RRT* (PIB-RRT*) and Potentially Guided Bi-directional RRT* (PB-RRT*). The proposed algorithms greatly improve the convergence rate and have a more efficient memory utilization. Theoretical and experimental evaluation of the proposed algorithms have been made and compared to the latest state of the art motion planning algorithms under different challenging environmental conditions and have proven their remarkable improvement in efficiency and convergence rate.

*Keywords:* Motion planning, Sampling based planning algorithms, RRT*, Optimal path planning, Artificial Potential Fields, Bidirectional trees.

## 1. Introduction

Motion planning has been a major problem since late 1980s [1] and due to the rise in robotics becoming a part of our everyday life, the research in this field has even become a greater need. In the motion planning problem, given the initial and goal configuration of the robot, the objective is to plan from the initial state to the goal region while avoiding obstacles along the way. Motion planning finds its application in our everyday life, in fields such as smart cars [2], robotic surgery [3], aerial, underwater and amphibious robotics [2], humanoid robotics [4] and in countless others [5] [6] [7] [8]. As humans explore the outer-space more and more, motion planning in outer-space [9] is also a becoming a challenging task. With the advancement in micro chip and nano-technology motion planning finds its application in nano-robotics and in micro-flow control and automation of bio-molecular computation (MF-BMC) [10].

Due to such a comprehensive requirement of motion planning, many motion planning algorithms were developed as mentioned in [11] [12] [13] [14] [15] [16] [17] [18]. Motion planning algorithms are either complete which return a solution if one exists in finite time and reports a failure if a solution does not exist, or are not complete but assure resolution or probabilistic completeness [11]. Complete motion planning algorithms such as Visibility Graphs have been developed. But these algorithms require explicit representation of the configuration space. Such a representation requires a lot of computational power especially for higher degree of freedom systems which renders such algorithms inefficient for practical purposes. In case of resolution or probabilistic complete algorithms, these algorithms either first discretize the given configuration space and then apply graph-based searches or use random sampling in case of sampling based algorithms respectively [11]. Graph-based search algorithms up to their resolution, find the optimal solution if one exists. One such example is "exact road maps", which discretizes the given configuration space that in turn places a heavy computational burden in higher dimensions. Another resolution complete algorithm, Artificial Potential Fields (APF) [19] also exists. But it is only effective if its resolution parameters are finely tuned, and another issue with APF is that it is greedy i.e it performs pure exploitation. Although exploitation might help compute the path quickly in some situations but it also causes the robot to get stuck in the local minima in APF [19].

*Corresponding author
Email address: zaid.tahir@smme.nust.edu.pk (Zaid Tahir[1])

Hence in order to avoid discretization of the state space and to avoid the above stated problems such as the local minima, sampling based stochastic searches were introduced such as Expansive Space Trees (ESTs) [20], Probabilistic Road Maps (PRMs) [21] and Rapidly-exploring Random Trees (RRTs) [22]. These stochastic sampling based searches are *probabilistically complete*, meaning that probability of finding a solution goes to one as the number of samples approach infinity. These algorithms performed very well in high dimensional spaces as well due to the fact that these algorithms do not require the complete construction of obstacle space. A major limitation of the above stated sampling-based algorithms is that they did not take into account the path cost and hence could not guarantee an optimal solution [23] [24]. Urmson and Simmons used heuristic-based sampling named h-RRT, to improve the path cost of RRT [25]. Ferguson et al. [26] used the anytime version of RRT, the anytime-RRTs improved the cost iteratively after finding an initial solution path. However, both these h-RRTs and anytime-RRTs did not guarantee optimal solution. Recently Karaman et al. [27] introduced an optimal variant of RRT, the RRT*. The RRT* first finds an initial path quickly, it then improves the solution by re-wiring the samples and replacing old parents with new parents whose cost in terms of Euclidean distance from initial state is less than them. This makes RRT* *asymptotically optimal*, meaning it guarantees convergence to optimal solution as the number of samples go to infinity. RRT* performs pure exploration, which can cause it to have very slow rates of convergence to optimal solution in highly cluttered environments and high dimensional spaces.

In order to improve the rates of convergence to optimal solution of RRT*, techniques such as sample-biasing [28] [29] [30], sample-rejection [31], sampling-heuristics [25], multiple trees [32], iterative searches [31] and anytime searches [26] were used. Qureshi et al. [29] used potential biasing on the randomly sampled states in RRT* to get to the optimal solution faster in his P-RRT* algorithm, which is an extension of previously proposed APGD-RRT* algorithm [33]. Karaman et al. [31] implemented the anytime version of RRT* using graph pruning. Multiple tree search based methods such as Bi-directional RRT* (B-RRT*) [32] and Intelligent Bi-directional RRT* (IB-RRT* ) [34] have recently been introduced and have shown to increase the convergence rate to optimal solution. In such a bidirectional search one Rapidly-exploring Random Tree (RRT) is grown from the initial state and the other Rapidly-exploring Random Tree is grown from one of the goal states. The bidirectional nature of these algorithms makes them inherently faster than the single-tree versions due to the fact that the samples that are too far away from the initial starting state are closer to the goal state. Whereas these samples are not directly connectable to the nodes of the growing tree in the single-tree version and could be used efficiently by connecting to the tree growing from the goal region in the bidirectional tree version. However, the problem with these bidirectional variants of RRT* is that they perform pure exploration, even though two trees are involved but there exist no sample-biasing to guide the two trees towards each other for faster convergence to optimal solution. In B-RRT* pure exploration is performed and each tree is grown one by one and a hybrid greedy connection heuristic checks if a connection between the two trees is possible or not. While IB-RRT* also performs pure exploration, a simple sampling heuristic adds the randomly sampled state to that tree out of the two bidirectional trees, from which the cost (in terms of Euclidean distance) of an obstacle free path from that tree to the randomly sampled state is less than the other tree. Then a connection heuristic checks if the two trees are connectable. Since, B-RRT* and IB-RRT* perform pure exploration, they also suffer in highly cluttered environments. In this paper, we introduce the concept of potentially guiding two Rapidly-exploring Random Trees towards each other in bidirectional sampling based motion planning by incorporating the proposed bidirectional potential gradient heuristics for alternatively directing each successive randomly sampled state towards each of the two trees out of the bidi-rectional trees and hence guiding both trees towards each other for faster convergence to optimal solution. This paper presents new bidirectional potential gradient heuristics, to potentially guide and directionalize two Rapidly-exploring Random trees towards each other in the bidirectional versions of RRT*, and hence the two proposed algorithms are the Potentially Guided Bi-directional RRT* (PB-RRT*) and Potentially Guided Intelligent Bi-directional RRT* (PIB-RRT*). The idea of potentially guiding two Rapidly exploring Random Trees towards each other for faster rate of convergence to optimal solution, as per authors knowledge, is novel.

In this paper PB-RRT* and PIB-RRT* have been rigorously tested in challenging 2-D and 3-D environments and are compared with the latest optimal sampling based algorithms such as RRT*, IB-RRT* and P-RRT*. The remainder of the paper is divided into following sections. Section **2** and **3** gives an explanation of our problem definition and a review of some previous algorithms. Section **4** explains our proposed algorithms, PB-RRT* and PIB-RRT*. Section **5** presents in depth analysis of the proposed PB-RRT* and PIB-RRT* algorithms regarding their probabilistic completeness, asymptotic optimality, rapid asymptotic convergence to optimal path, computational complexity and efficiency. Section **6** follows up with experimental proof, supporting the theoretical implications. Section **7** concludes the paper with some suggestions for further research in the future. Section **8** closes the paper with acknowledgments followed by references.

## 2. Problem definition

This section describes the motion planning algorithms that will be addressed in this paper along with the notations used. In the motion planning problem, a feasible path from initial state to the goal region has to be found in the least amount of time possible. Let $T_a$ and $T_b$ represent the two Rapidly-exploring Random Trees growing from initial and goal state, respectively. The state space of the configuration space is represented by the set $Z \subset \mathbb{R}^n$, $n \in \mathbb{N}$ and $n \geq 2$, where $n$ is the number of dimensions and $z \in Z$ is a particular configuration of the robot. $Z_{\text{obs}} \subset Z$ are the states that are present in our obstacle configuration space and are a no go configuration for the robot. $Z_{\text{free}}$ are the traversable states for the robot such that $Z_{\text{free}} = Z/Z_{\text{obs}}$. Let $V_a$ and $E_a$ be the vertices and edges of the tree $T_a$ such that $T_a = (V_a, E_a) \subset Z_{\text{free}}$. Similarly for tree $T_b$, $T_b = (V_b, E_b) \subset Z_{\text{free}}$. Let $\mu(.)$ be the Lebesgue measure, which denotes the $n$-dimensional volume of the given state space. Let a path be denoted by $\tau : [0, 1]$ and $\Sigma_{\text{free}}$ is the set of all collision free paths. Let $\tau_a'$ be the path of tree $T_a$ from initial state $z_{\text{init}}$ to any random state $z$, such that $\tau_a'[0, 1] \to \{\tau_a'(0) = z_{\text{init}}, \tau_a'(1) = z\} \subset Z_{\text{free}}$. Similarly for tree $T_b$, $\tau_b'$ is the path from goal state $z_{\text{goal}}$ to any random state $z$ such that $\tau_b'[0, 1] \to \{\tau_b'(1) = z, \tau_b'(0) = z_{\text{goal}}\} \subset Z_{\text{free}}$. In order to get a solution, both trees $T_a$ and $T_b$ must be connected such that $\tau_a'(1) = \tau_b'(1) = z$. Then the resulting concatenated solution will be given by $\tau_f' = \tau_a' \mid \tau_b' \in Z_{\text{free}}$. Finally, let $J(\tau)$ be the cost of the path $\tau$ in terms of Euclidean metric in $Z$. $U : \mathbb{R}^d \to \mathbb{R}$ describes the artificial potential function. Let $I_z$ describe the intensity of near vertices and $\vartheta_{\text{ALG}}$ denote the total rewiring per iteration of Algorithm ALG. Optimal path planning is a very basic requirement of motion planning. It is formally defined below.

**Optimal Path Planning:** Optimal path planning problem is formally defined as given a path planning triplet $\{z_{\text{init}}, Z_{\text{goal}}, Z_{\text{free}}\}$ and a path cost function $J(.)$. From all the feasible collision-free paths $\Sigma_{\text{free}}$, find a path $\tau^* \in \Sigma_{\text{free}}$ that minimizes the given function of path cost $J : \Sigma_{\text{free}} \to \mathbb{R} \geq 0$, such that $\tau^* : [0, 1] \to \tau^*(0) = z_{\text{init}}, \tau^*(1) = Z_{\text{goal}}$. Optimal path $\tau^*$ can hence be formally written as the following.

$$J(\tau^*) = \underset{\tau \in \Sigma_{\text{free}}}{\operatorname{argmin}} \{J(\tau) | \tau(0) = z_{\text{init}}, \tau(1) = Z_{\text{goal}}, \tau : [0, 1] \in \Sigma_{\text{free}}\}$$

## 3. Related work

In this section, previously proposed algorithms such as Potential Function Based-RRT* (P-RRT*) [29], Bi-directional RRT* (B-RRT*) [32] and Intelligent Bi-directional RRT* (IB-RRT*) [34] are briefly explained. These algorithms form the base of our proposed Potentially Guided Bidirectionalized RRT*.

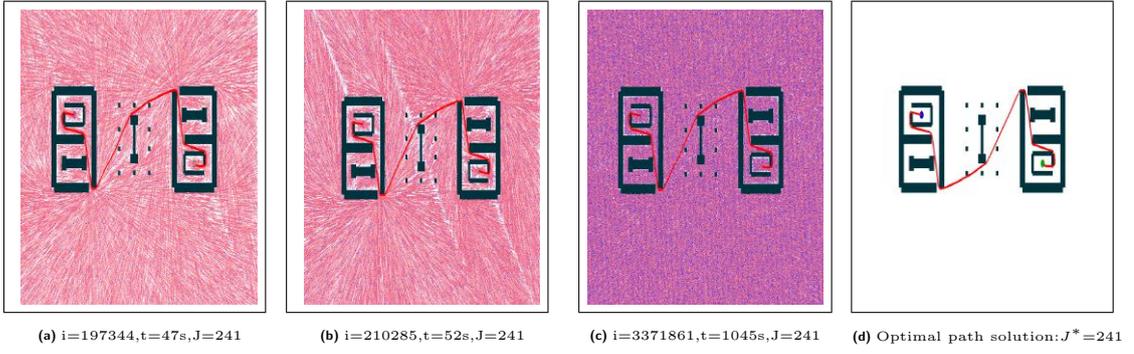

(a) i=197344,t=47s,J=241   (b) i=210285,t=52s,J=241   (c) i=3371861,t=1045s,J=241   (d) Optimal path solution:$J^*$=241

**Figure 1:** PIB-RRT*, PB-RRT* & RRT* performance comparison in 2-D Maze

---

**Algorithm 1:** P-RRT* ($z_{\text{init}}, z_{\text{goal}}$)
---
1   $V_a \leftarrow \{z_{\text{init}}\}$;
2   $E_a \leftarrow \emptyset$;
3   $T \leftarrow (V, E)$;
4   **for** $i \leftarrow 0$ **to** $N$ **do**
5      $z_{\text{rand}} \leftarrow \text{RandSample}(i)$;
6      $z_{\text{prand}} \leftarrow \text{RGD}(z_{\text{rand}})$;
7      $Z_{\text{near}} \leftarrow \text{NeighboringVertices}(i, T, z_{\text{prand}})$;
8      **if** $Z_{\text{near}} = \emptyset$ **then**
9         $Z_{\text{near}} \leftarrow \text{NearestVertex}(T, z_{\text{prand}})$;
10     $L \leftarrow \text{ListSorting}(z_{\text{prand}}, Z_{\text{near}})$;
11     $z_{\text{parent}} \leftarrow \text{PickBestParent}(L)$;
12     **if** $z_{\text{parent}}$ **then**
13        $T(V,E) \leftarrow \text{VertexInsert}(z_{\text{prand}}, z_{\text{parent}}, T)$;
14        $E \leftarrow \text{RewiringVertices}(z_{\text{prand}}, L, E)$;
15   **return** $T(V, E)$

---

### 3.1. P-RRT* Algorithm

The P-RRT* [29] Algoithm extends the RRT* Algorithm for better convergence properties by introducing a Random Gradient Descent (RGD) method. The RGD method guides the randomly sampled states towards the goal region using APF [19]. This guiding by P-RRT* results into faster computation of an optimal solution as compared to the original RRT* method. Algorithm 1 outlines the pseudo-code of P-RRT*. The procedures used in Algorithm 1 are described below.

---

**Algorithm 2:** ListSorting($z_{\text{rand}}, Z_{\text{near}}$)
---
1   $L \leftarrow \emptyset$;
2   **while** $z' \in Z_{\text{near}}$ **do**
3      $\tau' \leftarrow \text{Steer}(z', z_{\text{rand}})$;
4      $J' \leftarrow J(z_{\text{init}}, z') + J(z', z_{\text{rand}})$;
5      $L \leftarrow (z', J', \tau')$;
6   $L \leftarrow \text{Sort}(L)$;
7   **return** $L$

---

*RandSample:* This function returns independent and identically distributed (i.i.d.) samples form the obstacle-free space as $z_{\text{rand}} \in Z_{\text{free}}$.

*RGD:* This heuristic guides the randomly sampled state $z$ incrementally downhill in the direction of decreasing potential so that the resulting guided samples is $z_{\text{prand}}$.

*NeighboringVertices:* This procedure returns the vertices that are the neighboring vertices of the randomly sampled state $z_{\text{rand}}$ located inside a ball of volume $\mathcal{B}_{z,r}$ of radius $r$ centered at $z$ such that $r = \gamma(\log n/n)^{1/d}$, where $\gamma$ is a constant, $n$ is the number of vertices and $d$ is the dimension of the state space.

*NearestVertex:* As suggested from its name, this function returns the vertex from the tree $T(V, E)$ which is nearest to the randomly sampled vertex in terms of cost determined by the cost heuristic function $J()$.

*ListSorting:* This function sorts the list $L$ in terms of ascending order of its cost $J'$ as seen in Algorithm 2.

*Steer:* This function takes two states $z_1$ and $z_2$ as inputs and connects them in such a way that a straight trajectory collision-free path $\tau : [0, 1]$ is formed such that $\tau(0) = z_1$ and $\tau(1) = z_2$. Steering is done in small incremental step.

*PickBestParent:* This procedure chooses the parent $z_{\text{parent}} \in Z_{\text{near}}$ for the randomly sampled state which returns a collision-free path $\tau'$ with the minimum cost from $z_{\text{init}}$ to the randomly sampled state.

*VertexInsert:* Given a vertex $z$, this method assigns a parent $z_{\text{parent}}$ to the vertex $z$, and then computes the edge connecting the assigned parent $z_{\text{parent}}$ and the vertex $z$. Furthermore, it also determines the cost connecting $z_{\text{init}}$ to $z$ via $z_{\text{parent}}$. Finally, this vertex $z$ and edge are interested into the tree $T(V, E)$.

*RewiringVertices:* This function checks if the cost of the vertices in $Z_{\text{near}}$ in terms of Euclidean distance from the root state of the tree they belong to, is less through the new randomly sampled state than through their original parents. If so, then their original parents are removed and the randomly sampled vertex/state is made their new parent.

The P-RRT* Algorithm presented in Algorithm 1 is also asymptotically optimal like its baseline Algorithm RRT*. Along with being asymptotically optimal it is also probabilistically complete. But due to incorporation of APF into RRT* in P-RRT* it is many folds faster than its parent RRT* in finding the optimal solution especially in cluttered environments. First the tree $T(V, E)$ is initialized with its first vertex $z_{\text{init}}$. Then as iteration, $i$, goes from 0 to N, $z_{\text{rand}} \in Z_{\text{free}}$ is randomly sampled from obstacle-free configuration space $Z_{\text{free}}$. After this the randomly sampled state $z_{\text{rand}}$ is potentially guided by the $RGD()$ heuristic downhill towards decreasing potential so that it becomes $z_{\text{prand}}$ (line 6, Algorithm 1). Then the P-RRT* searches its neighboring vertices located around the potentially guided randomly sampled state $z_{\text{prand}}$ in a ball of of volume $\mathcal{B}_{z,r}$ to form the set $Z_{\text{near}}$. If this set $Z_{\text{near}}$ is empty it then populates $Z_{\text{near}}$ with the nearest vertex to $z_{\text{prand}}$ located anywhere in the tree (line 8–9, Algorithm 1). P-RRT* then makes a list $L$ of neighboring vertices $Z_{\text{near}}$ in the function ListSorting() as explained in Algorithm 2. This list $L$ is made in ascending order of the total cost $J'$. The total cost $J'$ (line 4, Algorithm 2) is the sum of the cost of a collision-free path (in terms of Euclidean distance) of the neighboring vertex $z' \in Z_{\text{near}}$ from root state $z_{\text{init}}$ plus the cost of the collision-free path from the neighboring vertex $z' \in Z_{\text{near}}$ to the potentially guided randomly sampled state $z_{\text{prand}}$. Then coming back to Algorithm 1, the function $PickBestParent()$ chooses the vertex $z' \in Z_{\text{near}}$ as the parent of the sample $z_{\text{prand}}$ which has the least cost $J'$ returning a collision-free path $\tau'$. Then the vertex $z_{\text{prand}}$ is added to the tree $T(V, E)$ and rewiring is done around $z_{\text{prand}}$ (line 12–14, Algorithm 1). This process is repeated until $i \to N$.

*3.2. B-RRT\* Algorithm*

Jordan and Perez [32] came up with optimal Bi-directional Rapidly-exploring Random trees in their B-RRT* Algorithm. It is implemented as shown in Algorithm 3. Some of the new functions used by B-RRT* are as follows.

*Extend:* Extend($z_1, z_2$) returns a new vertex $z_{\text{new}}$ such that $z_{\text{new}} \in Z_{\text{free}}$ and $z_{\text{new}}$ is closer to $z_2$ than $z_1$ in terms of cost, the Euclidean distance.

*Connect:* Algorithm 4 states the *Connect* heuristic in detail. Just like RRT-connect heuristic [35] it is greedy.

**Algorithm 3:** B-RRT* ($z_{\text{init}}, z_{\text{goal}}$)

1   $V_{\text{a}} \leftarrow \{z_{\text{init}}\}; V_{\text{b}} \leftarrow \{z_{\text{goal}}\};$
2   $E_{\text{a}} \leftarrow \emptyset; E_{\text{b}} \leftarrow \emptyset;$
3   $T_{\text{a}} \leftarrow (V_{\text{a}}, E_{\text{a}}); T_{\text{b}} \leftarrow (V_{\text{b}}, E_{\text{b}});$
4   $\tau_{\text{best}} \leftarrow \infty;$
5   **for** $i \leftarrow 0$ **to** $N$ **do**
6     $z_{\text{rand}} \leftarrow \text{RandSample}(i)$
7     $z_{\text{nearest}} \leftarrow \text{NearestVertex}(z_{\text{rand}}, T_{\text{a}})$
8     $z_{\text{new}} \leftarrow \text{Extend}(z_{\text{nearest}}, z_{\text{rand}})$
9     $Z_{\text{near}} \leftarrow \text{NeighboringVertices}(z_{\text{new}}, T_{\text{a}})$
10    $L \leftarrow \text{ListSorting}(z_{\text{new}}, Z_{\text{near}});$
11    $z_{\text{parent}} \leftarrow \text{PickBestParent}(L);$
12    **if** $z_{\text{parent}}$ **then**
13      $T_{\text{a}}(V_{\text{a}}, E_{\text{a}}) \leftarrow \text{VertexInsert}(z_{\text{new}}, z_{\text{parent}}, T_{\text{a}});$
14      $E_{\text{a}} \leftarrow \text{RewiringVertices}(z_{\text{new}}, L, E_{\text{a}});$
15    $z_{\text{conn}} \leftarrow \text{NearestVertex}(z_{\text{new}}, T_{\text{b}})$
16    $\tau_{\text{new}} \leftarrow \text{Connect}(z_{\text{new}}, z_{\text{conn}}, T_{\text{b}})$
17    **if** $\tau_{\text{new}} \neq \emptyset$ *and* $J(\tau_{\text{new}}) < J(\tau_{\text{best}})$ **then**
18      $\tau_{\text{best}} \leftarrow \tau_{\text{new}};$
19    $\text{SwapTrees}(T_{\text{a}}, T_{\text{b}});$
20   **return** $\{T_{\text{a}}, T_{\text{b}}\} = (V, E)$

---

**Algorithm 4:** Connect($z_1, z_2, T_{\text{b}}$)

1   $z_{\text{new}} \leftarrow Extend(z_2, z_1)$
2   $Z_{\text{near}} \leftarrow NeighboringVertices(z_{\text{new}}, T_{\text{b}})$
3   $List \leftarrow \text{ListSorting}(z_1, Z_{\text{near}});$
4   $z_{\text{parent}} \leftarrow \text{PickBestParent}(List);$
5   **if** $z_{\text{parent}}$ **then**
6     $E \leftarrow (z_{\text{parent}}, z_1);$
7     $\tau_{\text{free}} \leftarrow MakePath(z_{\text{parent}}, z_1);$
8     **return** $\tau_{\text{free}}$
9   **return** *NULL*

---

B-RRT* is explained in detail in Algorithm 3. It first initializes two trees, $T_{\text{a}}$ and $T_{\text{b}}$. $T_{\text{a}}$ is initialized by $z_{\text{init}}$ as its root vertex such that $z_{\text{init}} \in Z_{\text{free}}$. $T_{\text{b}}$ is initialized with $z_{\text{goal}}$ as its root vertex where $z_{\text{goal}} \in Z_{\text{goal}}$. The initial operations are just like RRT* where first a vertex is randomly sampled, then after insertion and rewiring of the sample into the selected tree $T_{\text{a}}$, $z_{\text{conn}}$ is searched, which is the nearest vertex from tree $T_{\text{b}}$ to the node $z_{\text{new}}$. Then the *Connect* heuristic tries to connect the two trees $T_{\text{a}}$ and $T_{\text{b}}$, returning a path $\tau_{\text{new}} \in \Sigma_{\text{free}}$. If the cost of the new path $J(\tau_{\text{new}})$ is less than the previously calculated best cost $J(\tau_{\text{best}})$, $\tau_{\text{best}}$ is overwritten by $\tau_{\text{new}}$. Then the trees $T_{\text{a}}$ and $T_{\text{b}}$ are swapped and the whole procedure is again executed until $i \rightarrow N$.

*3.3. IB-RRT* Algorithm*

Qureshi et al. [34] proposed their optimal bidirectional variant of RRT* in his Intelligent Bidirectional RRT* (IB-RRT*) Algorithm. The procedures used by this Algorithm are same as the ones used in RRT* except for the *GetBestTreeParent* heuristic, which has been explained below.

*GetBestTreeParent:* This heuristic calculates the best parent with the minimum cost from both trees $T_{\text{a}}$ and $T_{\text{b}}$ around the randomly sampled $z_{\text{rand}}$. Then the parent which has the least cost among the two trees $T_{\text{a}}$ and $T_{\text{b}}$ is made the parent of the randomly sampled state $z_{\text{rand}}$.

Algorithm 5 outlines the implementation of IB-RRT*. First the two trees $T_{\text{a}}$ and $T_{\text{b}}$ are initialized with their respective vertices (line 1–2). Then a vertex is randomly sampled and the near neighbor vertices from both trees are calculated in $Z^{\text{a}}_{\text{near}}$ and $Z^{\text{b}}_{\text{near}}$ (line 6–9). Both the neighboring vertex

**Algorithm 5:** IB-RRT* ($z_{\text{init}}, z_{\text{goal}}$)

1 $V_a \leftarrow \{z_{\text{init}}\}; V_b \leftarrow \{z_{\text{goal}}\}$;
2 $E_a \leftarrow \emptyset; E_b \leftarrow \emptyset$;
3 $T_a \leftarrow (V_a, E_a); T_b \leftarrow (V_b, E_b)$;
4 $\tau_{\text{best}} \leftarrow \infty$;
5 Connection $\leftarrow$ True
6 **for** $i \leftarrow 0$ **to** $N$ **do**
7 $\quad z_{\text{rand}} \leftarrow \text{RandSample}(i)$
8 $\quad \{Z_{\text{near}}^a, Z_{\text{near}}^b\} \leftarrow \text{NeghboringVertices}(z_{\text{rand}}, T_a, T_b)$
9 $\quad$ **if** $Z_{\text{near}}^a = \emptyset$ *and* $Z_{\text{near}}^b = \emptyset$ **then**
10 $\quad\quad \{Z_{\text{near}}^a, Z_{\text{near}}^b\} \leftarrow \text{NearestVertex}(z_{\text{rand}}, T_a, T_b)$
11 $\quad\quad$ Connection $\leftarrow$ False
12 $\quad L_a \leftarrow \text{ListSorting}(z_{\text{rand}}, Z_{\text{near}}^a)$
13 $\quad L_b \leftarrow \text{ListSorting}(z_{\text{rand}}, Z_{\text{near}}^b)$
14 $\quad \{z_{\text{parent}}, \text{flag}, \tau_{\text{free}}\} \leftarrow \text{GetBestTreeParent}(L_a, L_b, \text{Connection})$
15 $\quad$ **if** (flag) **then**
16 $\quad\quad T_a \leftarrow \text{VertexInsert}(z_{\text{rand}}, z_{\text{parent}}, T_a)$
17 $\quad\quad T_a \leftarrow \text{RewiringVertices}(z_{\text{rand}}, L_a, E_a)$
18 $\quad$ **else**
19 $\quad\quad T_b \leftarrow \text{VertexInsert}(z_{\text{rand}}, z_{\text{parent}}, T_b)$
20 $\quad\quad T_b \leftarrow \text{RewiringVertices}(z_{\text{rand}}, L_b, E_b)$
21 $E \leftarrow E_a \cup E_b$
22 $V \leftarrow V_a \cup V_b$
23 **return** $(\{T_a, T_b\} = V, E)$

sets $Z_{\text{near}}^a$ and $Z_{\text{near}}^b$ are sorted in ascending order by the function *ListSorting* (line 10–11). Then the *GetBestTreeParent* heuristic selects the best parent $z_{\min}$ from $T_a$ or $T_b$. If the best parent $z_{\min}$ is from $T_a$ then $z_{\text{rand}}$ is inserted in $T_a$ along with its edges and is then rewired (line 13–15). And if the best parent $z_{\min}$ is from $T_b$ then $z_{\text{rand}}$ is inserted in $T_b$ and rewired (line 16–18). This process continues till $i \rightarrow N$.

## 4. Potentially Guided Bidirectionalized RRT*

### 4.1. PB-RRT* & PIB-RRT*

In this section we present our proposed algorithms, PB-RRT* (Potentially Guided Bi-directional RRT*) and PIB-RRT* (Potentially Guided Intelligent Bi-directional RRT*). The proposed algorithms PB-RRT* and PIB-RRT* incorporate APF (Artificial Potential Fields) [19] into Bi-directional RRT* (B-RRT*) [32] and Intelligent Bi-directional RRT* (IB-RRT*) [34] respectively by using the proposed $BPG()$ (Bi-directional Potential Gradient) heuristic. $BPG()$ heuristic has been explained later in this section. The APF was introduced by Khatib [19]. In this method the artificial potential field $U_{\text{att}}$ pulls the robot $R_i$ located at position $z \in Z_{\text{free}}$ towards the goal region $z_g \in Z_{\text{goal}}$ and the artificial potential field $U_{\text{rep}}$ repels the robot away from obstacles lying in the obstacle configuration space $Z_{\text{obs}}$. Force $\hat{F}_r$ generated on the robot is the negative gradient of the resultant potential i.e, $\hat{F}_r = -grad[U_r]$.

$$U_{\text{att}} = \begin{cases} \frac{1}{2}k_p\|z-z_g\|^2, & \text{if } \|z-z_g\| > r_g \\ \frac{1}{2}k_p(r_g\|z-z_g\| - r_g^2), & \text{if } \|z-z_g\| \leq r_g \end{cases} \quad (1)$$

$$\hat{F}_{\text{att}} = \begin{cases} -k_p\|z-z_g\|, & \text{if } \|z-z_g\| > r_g \\ -k_p r_g \frac{\|z-z_g\|}{d(z,z_g)}, & \text{if } \|z-z_g\| \leq r_g \end{cases} \quad (2)$$

Where $r_g$ is the radius of the boundary around the goal region $z_g \in Z_{\text{goal}}$. $k_p$ is the attractive potential gain.

$$d_{\text{nearest}}^* = \underset{z' \in Z_{\text{obs}}}{\arg\min} \|z - z'\| \quad (3)$$

**Algorithm 6:** PB-RRT* ($z_{\text{init}}, z_{\text{goal}}$)

1   $V_a \leftarrow \{z_{\text{init}}\}; V_b \leftarrow \{z_{\text{goal}}\}$;
2   $E_a \leftarrow \emptyset; E_b \leftarrow \emptyset$;
3   $T_a \leftarrow (V_a, E_a); T_b \leftarrow (V_b, E_b)$;
4   $\tau_{\text{best}} \leftarrow \infty$;
5   **for** $i \leftarrow 0$ **to** $N$ **do**
6     $z_{\text{rand}} \leftarrow \text{RandSample}(i)$
7     $z_{\text{pb}} \leftarrow \text{BPG}(z_{\text{rand}}, i)$
8     $z_{\text{nearest}} \leftarrow \text{NearestVertex}(z_{\text{pb}}, T_a)$
9     $z_{\text{new}} \leftarrow \text{Extend}(z_{\text{nearest}}, z_{\text{pb}})$
10    $Z_{\text{near}} \leftarrow \text{NeighboringVertices}(z_{\text{new}}, T_a)$
11    $L \leftarrow \text{ListSorting}(z_{\text{new}}, Z_{\text{near}})$;
12    $z_{\text{parent}} \leftarrow \text{PickBestParent}(L)$;
13    **if** $z_{\text{parent}}$ **then**
14      $T_a(V_a, E_a) \leftarrow \text{VertexInsert}(z_{\text{new}}, z_{\text{parent}}, T_a)$;
15      $E_a \leftarrow \text{RewiringVertices}(z_{\text{new}}, L, E_a)$;
16    $z_{\text{conn}} \leftarrow \text{NearestVertex}(z_{\text{new}}, T_b)$
17    $\tau_{\text{new}} \leftarrow \text{Connect}(z_{\text{new}}, z_{\text{conn}}, T_b)$
18    **if** $\tau_{\text{new}} \neq \emptyset$ **and** $J(\tau_{\text{new}}) < J(\tau_{\text{best}})$ **then**
19      $\tau_{\text{best}} \leftarrow \tau_{\text{new}}$;
20    $\text{SwapTrees}(T_a, T_b)$;
21   **return** $\{T_a, T_b\} = (V, E)$

$$U_{\text{rep}} = \begin{cases} \frac{1}{2}k_{\text{rep}}(\frac{1}{d^*_{\text{nearest}}} - \frac{1}{d_{\text{obs}}})^2, & \text{if } d^*_{\text{nearest}} \leq d^*_{\text{obs}} \\ 0, & \text{if } d^*_{\text{nearest}} > d^*_{\text{obs}} \end{cases} \tag{4}$$

$$U_r = U_{\text{att}} + U_{\text{rep}} \tag{5}$$

$$\hat{F}_r = \hat{F}_{\text{att}} + \hat{F}_{\text{rep}} \tag{6}$$

In reference to Equation 3, $d^*_{\text{nearest}}$ is the distance of the robot to the nearest obstacle. $d_{\text{obs}}$ is a constant value and is usually very small. In Equation 4, $k_{\text{rep}}$ is the repulsive potential gain. The resultant force on the robot is $\hat{F}_r$ i.e, $\hat{F}_r = -grad[U_r]$ where $U_r = U_{\text{att}} + U_{\text{rep}}$ as seen in Equation 5. Force $F_r$ keeps acting on the robot until $-grad[U_r] = 0$ and when this zero potential gradient condition happens and $\hat{F}_r = 0$ then the robot has either reached its goal or is stuck in a local minima configuration.

The proposed PB-RRT* and PIB-RRT* fuse APF (Artificial Potential Fields) [19] with bi-directional variants of RRT* [32] [34] using the $BPG()$ heuristic. The pseudo code of PB-RRT* and PIB-RRT* is given in Algorithm 6 and 7 respectively. The flow of the algorithms PB-RRT* and PIB-RRT* is the same as that of B-RRT* (Algorithm 3) and IB-RRT* (Algorithm 4) respectively, with only difference of $BPG()$ heuristic. Hence only $BPG()$ heuristic is discussed in detail.

*4.2. BPG()*

APF [19] is incorporated in the proposed algorithms PB-RRT* and PIB-RRT* using the proposed $BPG()$ (Bi-directional Potential Gradient) heuristic. Pseudo-code of $BPG()$ heuristic is presented in Algorithm 8. Let $z_{\text{rand}} \in Z_{\text{free}}$ be the randomly sampled state. After being potentially guided by the $BPG()$ heuristic, the randomly sampled state $z_{\text{rand}}$ becomes potentially guided bidirectional randomly sampled state $z_{\text{pb}}$ such that $z_{\text{rand}} \rightarrow z_{\text{pb}}$, where $z_{\text{pb}} \in Z_{\text{free}}$. Some of the new heuristics used by $BPG()$ are discussed below.

*BPGgoal($Z_{\text{goal}}, z_{\text{pb}}$)*: $BPGgoal()$ (Bi-directional Potential Gradient towards goal state) implements Equation 1 and Equation 2 of the APF Algorithm in the form of Equation 7 and Equation 8 as shown below.

$$U_{\text{att}} = \frac{1}{2}k_p\|z_{\text{rand}} - z_{\text{goal}}\|^2 : z_{\text{goal}} \in Z_{\text{goal}} \tag{7}$$

$$\hat{F}_{\text{att}} = -k_p\|z_{\text{rand}} - z_{\text{goal}}\| : z_{\text{goal}} \in Z_{\text{goal}} \tag{8}$$

**Algorithm 7:** PIB-RRT* ($z_{\text{init}}, z_{\text{goal}}$)

1   $V_a \leftarrow \{z_{\text{init}}\}; V_b \leftarrow \{z_{\text{goal}}\}$;
2   $E_a \leftarrow \emptyset; E_b \leftarrow \emptyset$;
3   $T_a \leftarrow (V_a, E_a); T_b \leftarrow (V_b, E_b)$;
4   $\tau_{\text{best}} \leftarrow \infty$;
5   Connection $\leftarrow$ True
6   **for** $i \leftarrow 0$ **to** $N$ **do**
7       $z_{\text{rand}} \leftarrow$ RandSample($i$)
8       $z_{\text{pb}} \leftarrow$ BPG($z_{\text{rand}}, i$)
9       $\{Z_{\text{near}}^a, Z_{\text{near}}^b\} \leftarrow$ NeghboringVertices($z_{\text{pb}}, T_a, T_b$)
10      **if** $Z_{\text{near}}^a = \emptyset$ *and* $Z_{\text{near}}^b = \emptyset$ **then**
11          $\{Z_{\text{near}}^a, Z_{\text{near}}^b\} \leftarrow$ NearestVertex($z_{\text{pb}}, T_a, T_b$)
12         Connection $\leftarrow$ False
13      $L_a \leftarrow$ ListSorting($z_{\text{pb}}, Z_{\text{near}}^a$)
14      $L_b \leftarrow$ ListSorting($z_{\text{pb}}, Z_{\text{near}}^b$)
15      $\{z_{\text{parent}}, \text{flag}, \tau_{\text{free}}\} \leftarrow$ GetBestTreeParent($L_a, L_b$, Connection)
16      **if** (flag) **then**
17         $T_a \leftarrow$ VertexInsert($z_{\text{pb}}, z_{\text{parent}}, T_a$)
18         $T_a \leftarrow$ RewiringVertices($z_{\text{pb}}, L_a, E_a$)
19      **else**
20         $T_b \leftarrow$ VertexInsert($z_{\text{pb}}, z_{\text{parent}}, T_b$)
21         $T_b \leftarrow$ RewiringVertices($z_{\text{pb}}, L_b, E_b$)
22   $E \leftarrow E_a \cup E_b$
23   $V \leftarrow V_a \cup V_b$
24   **return** ($\{T_a, T_b\} = V, E$)

As seen in Algorithm 8 in the $BPG()$ heuristic if the iteration count $i$ is even then the potentially guided bidirectional randomly sampled state $z_{\text{pb}}$ is passed to the $BPGgoal()$ heuristic. In the $BPGgoal()$ heuristic, Equation 8 is used for the calculation of attractive potential force vector $\hat{F}_{\text{att}}$. $\hat{F}_{\text{att}}$ is acting on $z_{\text{pb}}$ with the goal region $Z_{\text{goal}}$ acting as the attractive pole pulling $z_{\text{pb}}$ towards it. Hence the name $BPGgoal()$ is given to this function. $k_p$ is the attractive potential gain. It is to be noted that the canonical portion of Equation 1 and Equation 2 is ignored in their implementation in Equation 7 and Equation 8 of $BPGgoal()$ as the potential force is being applied on the randomly sampled state and not physically on the robot hence it does not need to be slowed down by the canonical portion of the equation for avoidance of over-shooting $Z_{\text{goal}}$.

*NearestObstacleSearch*($Z_{\text{obs}}, z_{\text{pb}}$): This heuristic computes the distance $d_{\text{nearest}}^*$ of the nearest obstacle in the obstacle-space from the bidirectional potential gradient randomly sampled state $z_{\text{pb}}$.

*BPGinit*($z_{\text{init}}, z_{\text{pb}}$): $BPGinit()$ (Bi-directional Potential Gradient towards initial state) uses the Equation 1 and Equation 2 of the APF Algorithm in their modified form in Equation 9 and Equation 10 as shown below.

$$U'_{\text{att}} = \frac{1}{2} k_p \|z_{\text{rand}} - z_{\text{init}}\|^2 : z_{\text{init}} \in Z_{\text{free}} \tag{9}$$

$$\hat{F}'_{\text{att}} = -k_p \|z_{\text{rand}} - z_{\text{init}}\| : z_{\text{init}} \in Z_{\text{free}} \tag{10}$$

When the iteration count $i$ is odd in $BPG()$ heuristic then the potentially guided bidirectional randomly sampled state $z_{\text{pb}}$ is passed to the $BPGinit()$ heuristic. Using Equation 9 and Equation 10, $BPGinit()$ heuristic computes the attractive potential force vector $\hat{F}'_{\text{att}}$ acting on $z_{\text{pb}}$ with the initial root state $z_{\text{init}}$ acting as the attractive pole for $z_{\text{pb}}$. Hence the name $BPGinit()$ is given to this heuristic. Similarly as in $BPFGoal()$ heuristic, the canonical portion of Equation 1 and Equation 2 are ignored. $k_p$ is the attractive potential gain.

Algorithm 8 explains in detail $BPG()$ (Bi-directional Potential Gradient) heuristic. The randomly sampled state $z_{\text{rand}}$ is first fed to potentially guided bidirectional randomly sampled state variable $z_{\text{pb}}$ (line 1). The iteration count $i$ being currently run in the main loop is taken in $BPG()$

**Algorithm 8:** BPG ($z_{\text{rand}}, i$)

1   $z_{\text{pb}} \leftarrow z_{\text{rand}}$;
2   **if** $i \bmod 2 = 0$ **then**
3      **for** $k \leftarrow 0$ *to* $n$ **do**
4          $\hat{F}_{\text{att}} = BPGgoal(Z_{\text{goal}}, z_{\text{pb}})$;
5          $d^*_{\text{nearest}} \leftarrow$ NearestObstacleSearch($Z_{\text{obs}}, z_{\text{pb}}$);
6          **if** $d^*_{\text{nearest}} \leq d^*_{\text{obs}}$ **then**
7             **return** $z_{\text{pb}}$
8          **else**
9             $z_{\text{pb}} \leftarrow z_{\text{pb}} + \epsilon \frac{\hat{F}_{\text{att}}}{\|\hat{F}_{\text{att}}\|}$

10 **else**
11      **for** $k \leftarrow 0$ *to* $n$ **do**
12          $\hat{F}'_{\text{att}} = BPGinit(z_{\text{init}}, z_{\text{pb}})$;
13          $d^*_{\text{nearest}} \leftarrow$ NearestObstacleSearch($Z_{\text{obs}}, z_{\text{pb}}$);
14          **if** $d^*_{\text{nearest}} \leq d^*_{\text{obs}}$ **then**
15             **return** $z_{\text{pb}}$
16          **else**
17             $z_{\text{pb}} \leftarrow z_{\text{pb}} + \epsilon \frac{\hat{F}'_{\text{att}}}{\|\hat{F}'_{\text{att}}\|}$

18 **return** $z_{\text{pb}}$

heuristic as an input. If the iteration count $i$ is even then $z_{\text{pb}}$ is passed to the $BPGgoal()$ heuristic which computes the potential force vector $\hat{F}_{\text{att}}$ acting on $z_{\text{pb}}$ with the goal region $Z_{\text{goal}}$ acting as the attractive pole for $z_{\text{pb}}$ (line 4). Then the distance to the nearest obstacle $d^*_{\text{nearest}}$ from the sample $z_{\text{pb}}$ is computed (line 5). If this distance $d^*_{\text{nearest}}$ is smaller than a certain constant value $d^*_{\text{obs}}$ then the loop breaks and returns $z_{\text{pb}}$. $d^*_{\text{obs}}$ must be a very small value, its importance will be told in the coming section. But if $d^*_{\text{nearest}} > d^*_{\text{obs}}$ then the sample $z_{\text{pb}}$ is directed down-hill in direction of decreasing potential towards goal region in $\epsilon$ sized small steps (line 9). This loop continues in the same way until $k \to n$ where $n \in \mathbb{N}$. But if $i$ is odd then the potentially guided bidirectional randomly sampled state $z_{\text{pb}}$ is passed to the $BPGinit()$ heuristic which computes the potential force vector $\hat{F}'_{\text{att}}$ acting on the potentially guided bidirectional randomly sampled state $z_{\text{pb}}$ with the initial root state $z_{\text{init}}$ acting as the attractive pole for $z_{\text{pb}}$ (line 12). The rest of the procedure is same as mentioned above until either $d^*_{\text{nearest}} \leq d^*_{\text{obs}}$ or until $k \to n$. In this way for even iterations $i$, the potentially guided bidirectional randomly sampled state $z_{\text{pb}}$ is potentially directed down-hill towards goal region $Z_{\text{goal}}$ by the potential force vector $\hat{F}_{\text{att}}$, bringing it closer to tree $T_{\text{b}}$ being grown from the goal region $Z_{\text{goal}}$. While for odd iterations $i$, the potentially guided bidirectional randomly sampled state $z_{\text{pb}}$ is pulled towards initial root state $z_{\text{init}}$ by the potential force vector $\hat{F}'_{\text{att}}$ where the tree $T_{\text{a}}$ was grown hence bringing close both the trees $T_{\text{a}}$ and $T_{\text{b}}$ faster by the application of $BPG()$ (Bi-directional Potential Gradient) heuristic, hence achieving faster rate of convergence to optimal path as shown in the following sections. In order to keep a balance between exploitation and exploration, the value of $n$ in $k \leftarrow 0$ to $n$ (line 3,11), has to be chosen such that it is not too high that too much exploitation occurs or is too low that no exploitation occurs almost. In the following section we will be analysing our proposed algorithms.

## 5. Analysis

*5.1. Probabilistic Completeness*

Let $ALG$ denote any Algorithm and $G_{\text{i}}$ denote the sampling-based tree search graph with $i$ total iterations. $V_{\text{i}}$ is the set of vertices of the tree generated by Algorithm $ALG$ in $G_{\text{i}}^{\text{ALG}}$.

**Probabilistic Completeness:** For an Algorithm $ALG$, it is probabilistically complete, if for any path planning problem triplet $\{Z_{\text{free}}, z_{\text{init}}, Z_{\text{goal}}\}$, as the total number of iterations $i$ go to infinity, the probability of finding a feasible solution path from initial to goal configuration goes to one.

RRT* also ensure probabilistic completeness (Karaman and Frazzoli 2011) [27] as formally stated below.

**Theorem 1**([27]) *RRT\* is a probabilistically complete Algorithm. For any robustly feasible path planning problem triplet $\{Z_{\text{free}}, z_{\text{init}}, Z_{\text{goal}}\}$, as the number of iterations approach infinity, the probability of finding a feasible solution approaches one.*

$$\lim_{i \to \infty} P(\{\exists \, z_{\text{goal}} \in V_i^{\text{RRT*}} \cap Z_{\text{goal}} \; in \; G_i^{\text{RRT*}}\}) = 1$$

Similarly the proposed algorithms PB-RRT* and PIB-RRT* ensure probabilistic completeness as stated in the following Theorem 2.

**Theorem 2** *For a given feasible path planning problem triplet $\{Z_{\text{free}}, z_{\text{init}}, Z_{\text{goal}}\}$, the probability of finding a feasible solution is as follows.*

$$\lim_{i \to \infty} P(\{\exists \, z_{\text{goal}} \in V_i^{\text{PB-RRT*}} \cap Z_{\text{goal}} \; in \; G_i^{\text{PB-RRT*}}\}) = 1$$
$$\lim_{i \to \infty} P(\{\exists \, z_{\text{goal}} \in V_i^{\text{PIB-RRT*}} \cap Z_{\text{goal}} \; in \; G_i^{\text{PIB-RRT*}}\}) = 1$$

*As the number of iterations $i$ approach infinity, probability of finding a feasible solution if one exists, goes to one.*

*Sketch of proof:* For the proof of the above theorem we will use the following three arguments: 1) The bidirectional trees generated by PB-RRT* and PIB-RRT* just as in RRT* are connected trees i.e, whenever a state is randomly sampled, it is connected to its nearest neighbor state within the particular tree which is selected to grow by the PB-RRT* and PIB-RRT* algorithms respectively; 2) By convention we have $V_o^{\text{RRT*}} = V_o^{\text{PB-RRT*}} = V_o^{\text{PIB-RRT*}} = z_{\text{init}}$, hence the random trees generated by PB-RRT* and PIB-RRT* have $z_{\text{init}}$ as one of its states just like RRT*; 3) PB-RRT* and PIB-RRT* direct randomly sampled states towards goal region $Z_{\text{goal}}$ and initial root state $z_{\text{init}}$ at every even and odd loop iteration $i$ respectively by using the $BPG()$ heuristic. Therefore in the proposed algorithms PB-RRT* and PIB-RRT*, the probability that the two bi-directional trees grown with their roots at $z_{\text{init}}$ and $Z_{\text{goal}}$ respectively, will connect to each other and hence find a feasible path approaches to one as the number of iterations $i$ approach infinity. Based on the above stated arguments, it can be stated that the proposed algorithms PB-RRT* and PIB-RRT* ensure *Probabilistic Completeness*.

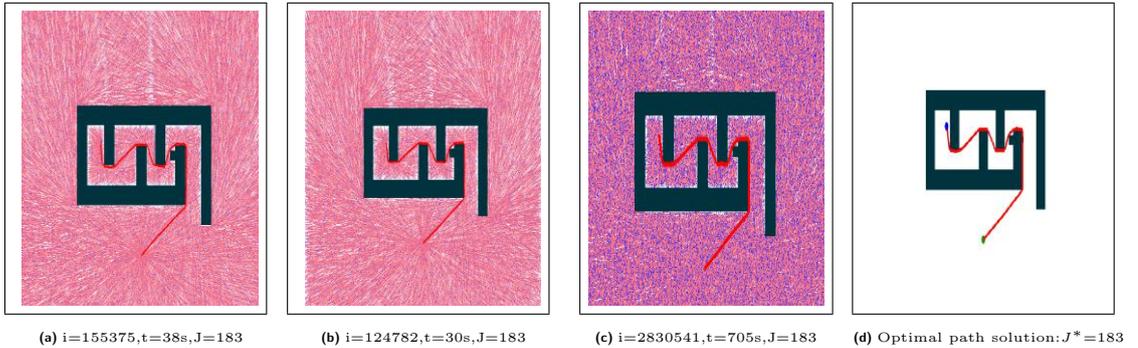

(a) i=155375,t=38s,J=183    (b) i=124782,t=30s,J=183    (c) i=2830541,t=705s,J=183    (d) Optimal path solution:$J^*$=183

**Figure 2:** PIB-RRT*, PB-RRT* & RRT* performance comparison in 2-D Box

*5.2. Asymptotic Optimality*

Let $\tau^* \in Z_{\text{free}}$ denotes the optimal path such that for a sequence of feasible paths $\{\tau_n\}$ where $\{\tau_n\} \in Z_{\text{free}} \; \forall n \in \mathbb{N}$ such that $\lim_{n \to \infty} \tau_n = \tau^*$ and $\lim_{n \to \infty} J(\tau_n) = J(\tau^*) = J^*$, where $J^*$ is the optimal path cost. An Algorithm is asymptotically optimal if it computes a minimum cost feasible path $\tau^* : [0,1] \to \tau^*(0) = z_{\text{init}}, \tau^*(1) = Z_{\text{goal}}$ and the optimal cost $J^*$ of the optimal path $\tau^*$ is the minimum achievable cost of any feasible path in the particular path planning problem $\{Z_{\text{free}}, z_{\text{init}}, Z_{\text{goal}}\}$. Let $Y_n^{\text{ALG}}$ be the extended random variable which corresponds to the minimum-cost solution returned in graph $G_n^{\text{ALG}}$ by the Algorithm $ALG$ at iteration $n$. Asymptotic optimality

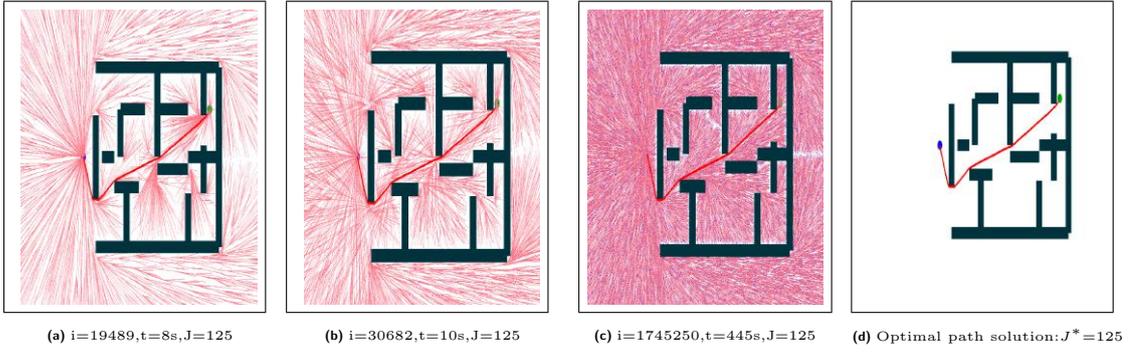

**(a)** i=19489,t=8s,J=125  **(b)** i=30682,t=10s,J=125  **(c)** i=1745250,t=445s,J=125  **(d)** Optimal path solution:$J^*$=125

**Figure 3:** PIB-RRT*, PB-RRT* & RRT* performance comparison in 2-D Cluttered

has been formally defined below.

**Asymptotic Optimality:** An Algorithm *ALG* is asymptotically optimal if for a cost function $J : \Sigma \to \mathbb{R} \geq 0$ it admits a robustly feasible solution with cost $J^*$.

$$P(\{\limsup_{n \to \infty} Y_n^{\text{ALG}} = J^*\}) = 1$$

**Theorem 3** *Let the conditions mentioned above in the definition of asymptotic optimality hold, then the proposed algorithms PB-RRT* and PIB-RRT* are asymptotically optimal if $d \geq 2$ and $\gamma > \gamma^* := 2^d(1 + \frac{1}{d})\mu(Z_{\text{free}})$.*

*Sketch of proof:* In reference to theorem 38 (Karaman and Frazzoli 2011) [27], PB-RRT* and PIB-RRT* our proposed potentially guiding bi-directional variants of RRT*, after potentially guiding the randomly sampled state $z_{\text{rand}} \in Z_{\text{free}}$, attempt to add nearby vertices in a radius of $r_n^{\text{PB-RRT*}} = \gamma^{\text{PB-RRT*}}(\frac{\log n}{n})^{\frac{1}{d}}$ and $r_n^{\text{PIB-RRT*}} = \gamma^{\text{PIB-RRT*}}(\frac{\log n}{n})^{\frac{1}{d}}$ respectively to form an edge so that the bi-directional trees are grown outwards in PB-RRT* and PIB-RRT*. As the same procedures happen is RRT*, therefore from Lemma 56, 71 and 72 (Karaman and Frazzoli 2011) [27] it can be derived that PB-RRT* and PIB-RRT* are asymptotically optimal as shown by the following derived relations.

$$P(\{\limsup_{n \to \infty} Y_n^{\text{PB-RRT*}} = J^*\}) = 1$$

$$P(\{\limsup_{n \to \infty} Y_n^{\text{PIB-RRT*}} = J^*\}) = 1$$

*5.3. Swift Convergence to Optimal Path*

Proof of swift convergence to optimal path of PB-RRT* and PIB-RRT* is given in this section based on the following assumptions.

**Assumption 1:** Let $\Sigma_{\text{free}}$ denote set of all collision free paths. Given two paths $\tau_1$ and $\tau_2$, let $\tau_1 \mid \tau_2$ denote their concatenation and $J(.)$ be the cost function such that for all $\tau_1, \tau_2 \in \Sigma_{\text{free}}$, $J(\tau_1) \leq J(\tau_1 \mid \tau_2)$.

**Assumption 2:** For $z \in Z_{\text{free}}$, there exists a ball of volume $\mathcal{B}_{z',\delta} \subset Z_{\text{free}}$ of radius $\delta \subset \mathbb{R} > 0$ centered around $z' \in Z_{\text{free}}$ such that $z' \in \mathcal{B}_{z',\delta}$.

Assumption 1 defines that if two paths are concatenated, then the combined cost will be no less than the individual cost of the paths. Assumption 2 tells us about the existence of collision free space around an obstacle and the path $\tau$ known as $\delta$-spacing which can be used to converge the path $\tau$ to optimal solution $\tau^*$. Having Assumption 2 under consideration let us define two terms, $\delta$-interior state $int_\delta(Z_{\text{free}})$ and $\delta$-exterior state $ext_\delta(Z_{\text{free}})$. If a ball region of volume $\mathcal{B}_{z,\delta}$ of radius $\delta$ centered at $z$ lies entirely inside the collision-free space $Z_{\text{free}}$, then $z$ is said to be in the $\delta$-interior state of $Z_{\text{free}}$. The $\delta$-interior of $Z_{\text{free}}$ is defined as $int_\delta(Z_{\text{free}}) := \{z \in Z_{\text{free}} \mid \mathcal{B}_{z,\delta} \subseteq Z_{\text{free}}\}$. While $\delta$-exterior states are those states that are close to the obstacles but not entirely inside, such that for a state $z' \in Z_{\text{free}}$ a ball region of volume $\mathcal{B}_{z',\delta}$ of radius $\delta$ centered at $z'$ lies partially inside the collision-free space $Z_{\text{free}}$ and some of the states in its volume lie in obstacle space $Z_{\text{obs}}$, then $z'$ is said to lie in the $\delta$-exterior state. The $\delta$-exterior state is represented as $ext_\delta(Z_{\text{free}}) := Z_{\text{free}}/int_\delta(Z_{\text{free}})$. Paths with a strong and a weak $\delta$-clearance will be explained below.

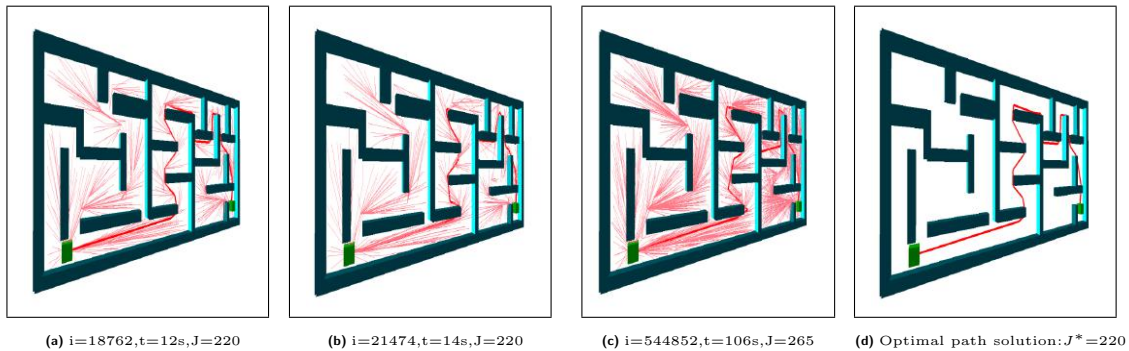

(a) i=18762,t=12s,J=220  (b) i=21474,t=14s,J=220  (c) i=544852,t=106s,J=265  (d) Optimal path solution:$J^*$=220

**Figure 4:** PIB-RRT*, PB-RRT* & RRT* performance comparison in 3-D Maze

*Path with strong $\delta$-clearance:* If a feasible path $\tau \in \Sigma_{\text{free}}$ comprises entirely of $\delta$-interior states of $Z_{\text{free}}$, then that path is said to have a strong $\delta$-clearance for $\delta > 0$.

*Path with weak $\delta$-clearance:* Let $\tau_1 : [0,1]$ and $\tau_2 : [0,1]$ be two collision free paths. Both paths have same initial and goal states $\tau_1(0) = \tau_2(0)$, $\tau_1(1) = \tau_2(1)$ respectively. The path $\tau_1$ is homotopic to the path $\tau_2$ by the homotopy function $\psi : [0,1]$ such that $\psi(q)$ is a collision free path where $q \to [0,1]$ and $\psi(0) = \tau_1$ has weak $\delta$-clearance and $\psi(1) = \tau_2$ has strong $\delta$-clearance and for all $\beta \in [0,1]$, their is $\delta_\beta > 0$ such that $\psi(\beta)$ has $\delta_\beta$-clearance.

The following lemma states that the proposed $BPG()$ heuristic potentially guides the random samples $z \in Z_{\text{free}}$ towards weak $\delta$-clearance region where optimal solution exists.

**Lemma 1** The $BPG()$ heuristic directs the random sample $z \in Z_{\text{free}}$ towards $\delta$-exterior $ext_\delta(Z_{\text{free}})$ region, where $\delta$-clearance is weak.

*Sketch of proof:* The $BPG()$ heuristic causes the random sample $z \in Z_{\text{free}}$ to be potentially pulled by the attractive pole which is either the goal region $Z_{\text{goal}}$ or the initial root state region $z_{\text{init}}$ which are swapped on each alternate iteration. As the random sample is being potentially guided down the slope towards the attracting pole in small $\epsilon$ steps for $n$ time or till a very minute distance $d^*_{\text{obs}}$ from any obstacle is reached as seen in Algorithm 8. The minute distance $d^*_{\text{obs}}$ causes the random sample to achieve weak $\delta$-clearance as it is being potentially guided there.

Let $\tau'_n$ and $\tau_n$ be two paths such that $\tau'_n \in \Sigma_{\text{free}}$ and $\tau_n \in \Sigma_{\text{free}}$ and $\tau'_n$ is closest to $\tau_n$ in terms of bounded variation norm among all paths in $\Sigma_{\text{free}}$. The following lemma states that convergence to optimal solution is surely guaranteed if, the random variable $\|\tau'_n - \tau_n\|_{\text{BV}}$ converges to zero

**Lemma 2**([4]) A sampling based Algorithm converges to optimal solution if, the random variable $\|\tau'_n - \tau_n\|_{\text{BV}}$ converges to zero i.e,

$$P(\{\lim_{n \to \infty} \|\tau'_n - \tau_n\|_{\text{BV}} = 0\}) = 1$$

**Corollary 2** As the number of iterations approach infinity $\tau'_n$ will eventually converge to the optimal path $\tau^*$.

$$P(\{\lim_{n \to \infty} \tau'_n = \tau^*\}) = 1$$

Let $I_z$ denote the intensity of near vertices $Z_{\text{near}}$ around a random state $z \in Z_{\text{free}}$ in a ball of radius $r$ such that

$$I_z := \{card(Z_{\text{near}}/\mu(\mathcal{B}_{z,r}) : Z_{\text{near}} \mid z = \mathcal{B}_{z,r} \cap V_n\}$$

*Sketch of proof:* Let $\varepsilon \in \mathbb{R}$. $Z_{\text{near}}$ is the set of near vertices around a randomly sampled state $z_{\text{rand}} \in Z_{\text{free}}$ located inside a ball of volume $\mathcal{B}_{z,r}$ of radius $r$ centered at $z$ such that $r = \gamma(\log n/n)^{1/d}$, where $\gamma$ is a constant, $n$ is the number of vertices and $d$ is the dimension of the state space. The randomly sampled state $z_{\text{rand}}$ can make any state $z' \in Z_{\text{near}}$ as its parent which has the lowest cost of connecting $z_{\text{rand}}$ to the root $z_{\text{init}}$ of the tree out of all the vertices in $Z_{\text{near}}$. This means that $\|z_{\text{rand}} - z'\| = \varepsilon$, where $\varepsilon < r = \gamma(\log n/n)^{1/d}$. This ensures that the tree in the algorithm grows in small incremental steps of $d'$ where $d' \leq \varepsilon$. For incremental expansion or wavefront expansion of trees it is proven that regions near the root of trees are more dense [22]. Hence if the randomly sampled state $z_{\text{rand}}$ lies closer to the generation of the tree, there is a high probability of having high cardinality of set $Z_{\text{near}}$.

The proposed algorithms PB-RRT* and PIB-RRT* are designed to converge to optimal solution very quickly. Since the intensity of near vertices $I_z$ is higher in region closer to the generation of the tree [22], following *Lemma 2* states that Bi-directional Potential Gradient $BPG()$ heuristic directs the random samples towards higher intensity $I_z$ regions where higher probability of optimal solution exists.

**Lemma 3** The proposed $BPG()$ heuristic guides the random sample $z \in Z_{\text{free}}$ towards the regions closer to point of generation of the trees where the intensity of near vertices $I_z$ is much higher.

*Sketch of proof:* The $BPG()$ heuristic of the proposed PB-RRT* and PIB-RRT* guides the random sample $z \in Z_{\text{free}}$ towards the goal region $Z_{\text{goal}}$ and initial root state $z_{\text{init}}$ as the poles of attraction change on every iteration in $BPG()$, directing the random sample down the slope under the influence of the attractive pole, either it be $z_{\text{init}}$ or $Z_{\text{goal}}$ region pole. These are the points of generation of the bidirectional trees $T_{\text{a}}$ and $T_{\text{b}}$, generated at $z_{\text{init}}$ and $Z_{\text{goal}}$ respectively. Hence this means that the intensity of near vertices $I_z$ is much higher at the root of the trees and the random sample $z \in Z_{\text{free}}$ is directed towards higher $I_z$ region causing increased rewiring per iteration of both proposed algorithms represented as $\vartheta_{\text{PB-RRT*}}$ and $\vartheta_{\text{PIB-RRT*}}$ respectively. Such potential fields in bidirectional tree search do not exist in RRT* hence, $\vartheta_{\text{PB-RRT*}} > \vartheta_{\text{RRT*}}$ and $\vartheta_{\text{PIB-RRT*}} > \vartheta_{\text{RRT*}}$. Since rewiring tries to minimize the bounded variation $\|\tau'_{\text{n}} - \tau_{\text{n}}\|_{\text{BV}}$ as mentioned before, causes convergence to optimal path faster than RRT*.

Based on Lemma 2 and Lemma 3, Theorem 4 is formalized as follows stating that the convergence to the optimal solution of the proposed PB-RRT* and PIB-RRT* is faster than RRT* due to increased rewiring per iterations in the proposed algorithms than RRT* as explained above.

**Theorem 4** *From Lemma 2 and Lemma 3 it is derived that $BPG()$ heuristic in PB-RRT* and PIB-RRT* guides the random sample $z \in Z_{\text{free}}$ towards higher intensity $I_z$ regions such that $\vartheta_{\text{PB-RRT*}} > \vartheta_{\text{RRT*}}$ and $\vartheta_{\text{PIB-RRT*}} > \vartheta_{\text{RRT*}}$.*

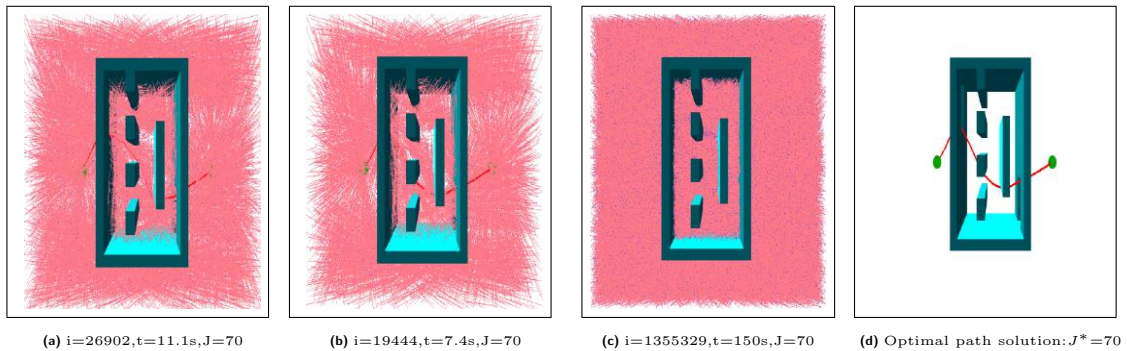

(a) i=26902,t=11.1s,J=70    (b) i=19444,t=7.4s,J=70    (c) i=1355329,t=150s,J=70    (d) Optimal path solution:$J^*$=70

**Figure 5:** PIB-RRT*, PB-RRT* & RRT* performance comparison in 3-D Columns

On the basis of Theorem 4, Lemma 4 has been derived and is stated as follows.

**Lemma 4** In the given path planning problem $\{Z_{\text{free}}, z_{\text{init}}, Z_{\text{goal}}\}$, the proposed $BPG()$ heuristic guides the random sample $z \in Z_{\text{free}}$ in such a manner that the two tree in the bi-directional search are potentially pulled towards each other.

*Sketch of proof:* The $BPG()$ heuristic contains $BPFgoal()$ and $BPFinit()$ heuristics. $BPFgoal()$ heuristic causes the goal region $Z_{\text{goal}}$ to become the attractive pole while $BPFinit()$ heuristic causes the initial root state region $z_{\text{init}}$ to become the attractive pole. As there are two trees $T_{\text{a}}$ and $T_{\text{b}}$ initial root state and goal region respectively in the proposed PB-RRT* and PIB-RRT*. $BPFgoal()$ and $BPFinit()$ potentially pull the random sample towards the attracting pole which is either the goal region where the origin of tree $T_b$ is located or towards $z_{\text{init}}$ which is the origin of tree $T_a$ and on the next iteration the pole is swapped. This causes the random sample $z \in Z_{\text{free}}$ to be potentially pulled towards origins of both trees, $T_a$ and $T_b$ on alternate iterations causing both trees to grow and be potentially pulled towards each other due to the $BPG()$ heuristic resulting faster convergence to optimal solution. Based on Corollary 2 and Lemmas 1,2,3 and 4, Theorem 5 formally states the faster convergence of PB-RRT* and PIB-RRT* due to the $BPG()$ heuristic.

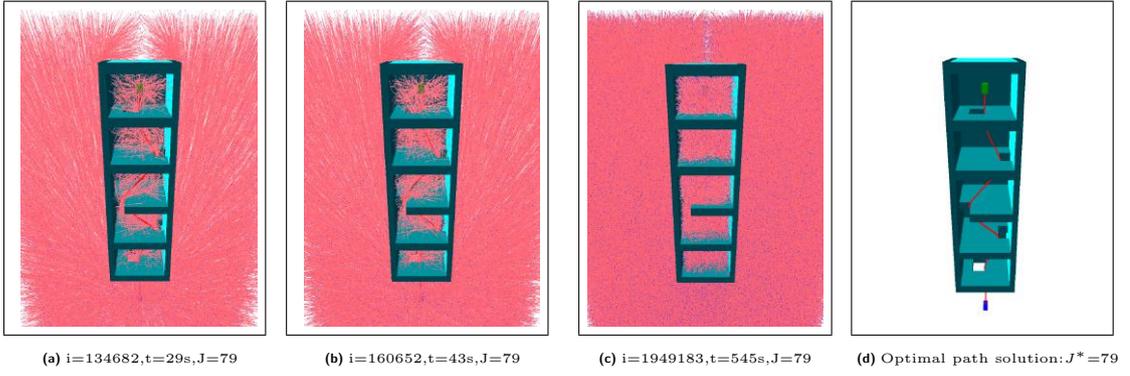

(a) i=134682,t=29s,J=79  (b) i=160652,t=43s,J=79  (c) i=1949183,t=545s,J=79  (d) Optimal path solution:$J^*$=79

**Figure 6:** PIB-RRT*, PB-RRT* & RRT* Initial performance comparison in 3-D Cluttered

**Theorem 5** *The BPG() (Bi-directional Potential Gradient) heuristic (**1**) potentially directs the random sample $z \in Z_{\text{free}}$ towards higher intensity $I_z$ regions where rewirings per iteration are related as, $\vartheta_{\text{PB-RRT*}} > \vartheta_{\text{RRT*}}$ and $\vartheta_{\text{PIB-RRT*}} > \vartheta_{\text{RRT*}}$; (**2**) the random sample after being potentially guided becomes $z_{\text{pb}}$ so that $P(z_{\text{pb}} \in Z_{\text{ext}_\delta}) > 0$; (**3**) The resulting solution path $\tau$ very quickly converges to optimal path $\tau^*$ so that $\|\tau'_n - \tau_n\| = 0$ where $\tau'_n = \tau^*$.*

Hence from Theorem 1,2,3,4 and 5 it has been deduced that the proposed PB-RRT* and PIB-RRT* algorithms find the feasible solution to a motion planning problem and converge to optimal solution very quickly.

*5.4. Computational Complexity*

The computational complexity of PB-RRT* and PIB-RRT* has been discussed in this section. Let $M_n^{\text{ALG}}$ define the total computations performed by Algorithm $ALG$. $M_n^{\text{PB-RRT*}}$ and $M_n^{\text{PIB-RRT*}}$ are the total processes performed by PB-RRT* and PIB-RRT* respectively. Theorem 6 proposes that the computational complexity of PB-RRT* and PIB-RRT* are a constant times higher than that of RRT* where as Theorems 7 and Theorem 8 state the comparison of PB-RRT* with B-RRT* and PIB-RRT* with IB-RRT* respectively.

**Theorem 6** *There exists constants $\Phi_1$, $\Phi_2 \in \mathbb{R}_+$ such that the computational complexity ratio of PB-RRT* and PIB-RRT* with RRT* is as follows.*

$$\limsup_{n \to \infty} \mathbb{E}\left[\frac{M_n^{\text{PB-RRT*}}}{M_n^{\text{RRT*}}}\right] \leq \Phi_1$$

$$\limsup_{n \to \infty} \mathbb{E}\left[\frac{M_n^{\text{PIB-RRT*}}}{M_n^{\text{RRT*}}}\right] \leq \Phi_2$$

**Theorem 7** *The computational complexity ratio of PB-RRT* and B-RRT* is such that there exists a constant $\Phi_3 \in \mathbb{R}_+$*

$$\limsup_{n \to \infty} \mathbb{E}\left[\frac{M_n^{\text{PB-RRT*}}}{M_n^{\text{B-RRT*}}}\right] \leq \Phi_3$$

**Theorem 8** *Let $\Phi_4 \in \mathbb{R}_+$ be a constant so that the computational complexity ratio of PIB-RRT* and IB-RRT* is as follows.*

$$\limsup_{n \to \infty} \mathbb{E}\left[\frac{M_n^{\text{PIB-RRT*}}}{M_n^{\text{IB-RRT*}}}\right] \leq \Phi_4.$$

*Sketch of proof:* When compared to RRT* along with being bi-directional another procedure $BPG()$ has been incorporated in the proposed PB-RRT* and PIB-RRT*. PB-RRT* has an additional $Connect()$ heuristic along with $BPG()$ heuristic. PIB-RRT* has $GetBestTreeParent()$ in place of $PickBestParent()$ in RRT* and $BPG()$ heuristic. $BPG()$ heuristic can be executed in a constant number of iterations and does not depend upon the number of vertices in the tree. It has to find nearest obstacle form the random sample $z \in Z_{\text{free}}$ which requires at least $\Omega(\log_{10} n)$ time. Furthermore in PB-RRT* and PIB-RRT* both execute $NearestVertex()$ and $NeighbouringVertices()$ procedure for both trees $T_a$ and $T_b$ just like RRT* which adds up a constant computation overhead when compared to RRT* in $\log_{10} n$ terms. Hence as seen in Theorem 6, PB-RRT* and PIB-RRT* only vary from RRT* by $\Phi_1$ and $\Phi_2$ in terms of computational complexity ratio.

In Theorem 7 the computational complexity ratio of PB-RRT* and B-RRT* is given and as $BPG()$ heuristic is the only additional procedure in PB-RRT* as compared to B-RRT*, hence their computational complexity only differs by a constant ratio $\Phi_3 \in \mathbb{R}_+$. Similarly in Theorem 8 PIB-RRT* differs from IB-RRT* by a constant $\Phi_4 \in \mathbb{R}_+$ in computational complexity ratio.

| Environment | Algorithm | $i_{\min}$ | $i_{\max}$ | $i_{\text{avg}}$ | $t_{\min}(s)$ | $t_{\max}(s)$ | $t_{\text{avg}}(s)$ | $\vartheta_{\text{avg}}$ | $J(\tau*)$ | Fail |
|---|---|---|---|---|---|---|---|---|---|---|
| 2D-Maze (Fig 1) | PIB-RRT* | 180,484 | 224,306 | 198,608 | 44 | 57 | 48 | 1.09 | 241 | |
| | PB-RRT* | 191,273 | 234,547 | 215,505 | 47 | 58.7 | 53 | 1.02 | 241 | |
| | IB-RRT* | 256,374 | 335,877 | 290,126 | 62 | 70 | 66 | 0.74 | 241 | |
| | P-RRT* | 300,001 | 375,661 | 332,127 | 67 | 74 | 72 | 0.66 | 241 | |
| | RRT* | 3,234,592 | 3,799,518 | 3,371,861 | 922 | 1140 | 1045 | 0.35 | 241 | 70% |
| 2D-Box (Fig 2) | PIB-RRT* | 122,355 | 189,826 | 157,941 | 34 | 43 | 38 | 0.84 | 183 | |
| | PB-RRT* | 100,001 | 158,870 | 125,186 | 25 | 39 | 30.4 | 0.93 | 183 | |
| | IB-RRT* | 214,341 | 271,247 | 232,315 | 54 | 61 | 59 | 0.70 | 183 | |
| | P-RRT* | 247,984 | 330,541 | 294,187 | 61 | 71 | 66 | 0.65 | 183 | |
| | RRT* | 2,247,984 | 2,830,541 | 2,424,187 | 611 | 705 | 687 | 0.31 | 183 | 50% |
| 2D-Cluttered (Fig 3) | PIB-RRT* | 13,452 | 26,731 | 20,539 | 6.3 | 9.9 | 8.1 | 1.25 | 125 | |
| | PB-RRT* | 17,537 | 31,345 | 22,124 | 7.2 | 10.1 | 8.7 | 1.18 | 125 | |
| | IB-RRT* | 35,164 | 39,402 | 38,152 | 12 | 14 | 13.5 | 0.97 | 125 | |
| | P-RRT* | 66,373 | 71,324 | 68,332 | 24 | 28 | 26.5 | 0.63 | 125 | |
| | RRT* | 1,458,373 | 1,926,184 | 1,670,140 | 320 | 531 | 420 | 0.39 | 125 | 50% |
| 3D-Maze (Fig 4) | PIB-RRT* | 14,011 | 19,719 | 17,430 | 10.5 | 13 | 11.5 | 0.98 | 220 | |
| | PB-RRT* | 18,602 | 24,361 | 21,459 | 11.7 | 15 | 14.1 | 0.88 | 220 | |
| | IB-RRT* | 33,400 | 42,100 | 38,213 | 18.2 | 22 | 19.5 | 0.77 | 220 | |
| | P-RRT* | 74,300 | 79,613 | 77,641 | 28 | 31.5 | 29.6 | 0.62 | 220 | |
| | RRT* | - | - | - | - | - | - | - | - | 100% |
| 3D-Columns (Fig 5) | PIB-RRT* | 23,484 | 26,902 | 24,629 | 9.3 | 11.1 | 10.4 | 0.99 | 70 | |
| | PB-RRT* | 17,624 | 21,337 | 19,647 | 6.2 | 8.1 | 7.5 | 1.05 | 70 | |
| | IB-RRT* | 35,444 | 43,404 | 39,583 | 14 | 21 | 18 | 0.71 | 70 | |
| | P-RRT* | 81,561 | 84,321 | 83,523 | 32.5 | 34 | 33 | 0.58 | 70 | |
| | RRT* | 678,001 | 1,246,149 | 900,132 | 124 | 271 | 174 | 0.43 | 70 | 70% |
| 3D-Cluttered (Fig 6) | PIB-RRT* | 110,629 | 151,784 | 131,542 | 26.3 | 32 | 28.5 | 1.31 | 79 | |
| | PB-RRT* | 130,242 | 177,804 | 153,358 | 31 | 47 | 41 | 1.12 | 79 | |
| | IB-RRT* | 200,963 | 244,618 | 223,493 | 59 | 64 | 61 | 1.02 | 79 | |
| | P-RRT* | 352,693 | 385,284 | 369,381 | 72 | 75 | 74 | 0.91 | 79 | |
| | RRT* | 1,857,381 | 1,949,183 | 1,908,392 | 514 | 545 | 529 | 0.63 | 79 | 60% |

## 6. Experimental Results

In this section the numerical path planning experiments are provided that make a comparison of the proposed PB-RRT* and PIB-RRT* algorithms with other aysmptotically optimal sampling based algorithms such as IB-RRT*, P-RRT* and RRT* on a variety of environments. Different state dimensions were used in different environments to thoroughly test the algorithms. The size and configuration space of the different environments was varied from very large to very small but was kept constant for a particular environment so that a just comparison could be made between different algorithms while simulating on that environment. Due to randomization the algorithms were run up to 50 times with a common pseudo-random seed which was varied on every re-run of the Algorithm. The simulations were performed on a $2.30GHz$ Intel core $i3$ processor with $2GB$ RAM.

If the total number of iterations exceeded $5 \times 10^6$, the simulation was terminated and the result was declared as failed to restrict the computational time. The results of the above mentioned algorithms in different environments are provided in Table 1. Maximum, minimum and average iterations and time are provided in Table 1. Fail columns denote the percentage of the total 50 reruns the Algorithm has failed to find the optimal solution. The column $\vartheta_{\text{avg}}$ rewiring per iteration

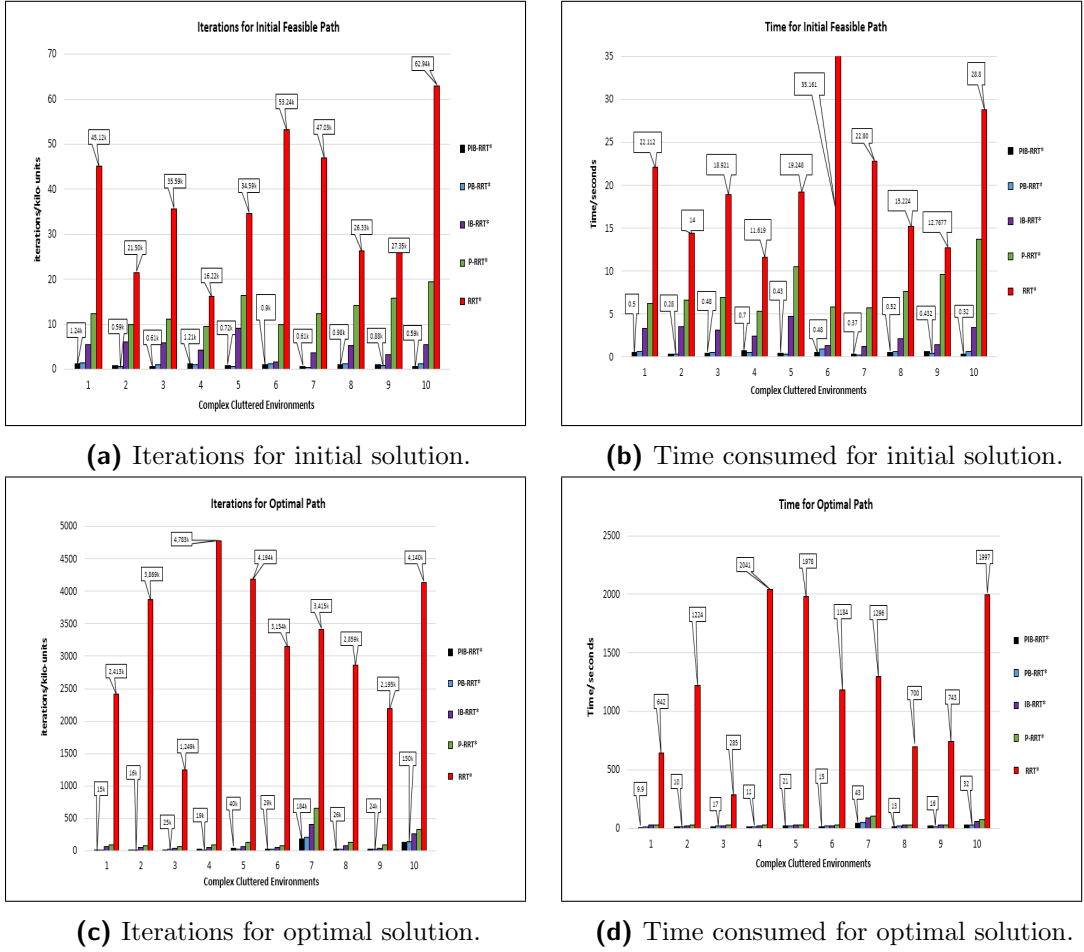

**(a)** Iterations for initial solution.

**(b)** Time consumed for initial solution.

**(c)** Iterations for optimal solution.

**(d)** Time consumed for optimal solution.

**Figure 7:** Performance Comparison in 10 complex cluttered environments

denotes the average number of rewiring occurring from start to termination of the simulation. $J^*$ indicates the optimal path cost in terms of Euclidean metric of the particular environment within a specified tolerance of the true optimum. PB-RRT* and PIB-RRT* along with the above mentioned algorithms were tested in a variety of environments (e.g, Figs 1-6), though only the figures of simulations of PB-RRT*, PIB-RRT* and RRT* are displayed due to limited space available. Each environment tested a different aspect of the Algorithm. Results of the testing and simulations are discussed below.

Fig 1 is a 2-D highly cluttered environment used to test the efficiency of the above mentioned algorithms in an environment ladened with obstacles, where it is very hard to find the optimal solution. As seen from Table 1, PIB-RRT* is able to find the optimal solution in this highly cluttered environment the most quickly ($i_{\text{avg}} = 198,608$) with highest rewiring per iteration average ($\vartheta_{\text{PIB-RRT*}} = 1.09$). After PIB-RRT*, PB-RRT*, IB-RRT* and P-RRT* find the optimal solution the quickest respectively while $RRT*$ failed to find the optimal solution 70% of the times within the specified range. PIB-RRT* is quickest to find the optimal solution due to the *Intelligent Sample Insertion* it inherited from IB-RRT* which was specifically designed for highly cluttered environments and coupling *Intelligent Sample Insertion* with the proposed $BPG()$ heuristic for PIB-RRT* gave us the optimal solution in the least amount of time. After PIB-RRT*, PB-RRT* was the quickest in finding the optimal solution of Fig 1 due to the $BPG()$ heuristic coupled with Bi-directional RRT* (B-RRT*). Fig 2 is another 2-D environment, as seen it is not as highly cluttered as Fig 1 and PB-RRT* is the quickest to find the optimal solution in this environment due to the $BPG()$ heuristic pulling both Bi-directional trees towards each other, coupled with the partial greedy heuristic of B-RRT* it has given the optimal solution in this environment the most quickly taking least number of average iterations to optimal solution at ($i_{\text{avg}} = 125,186$). After PB-RRT*, PIB-RRT* was second in least number of iterations to optimal solution at ($i_{\text{avg}} = 157,941$). Then came IB-RRT* and P-RRT* respectively at higher average iterations to optimal solution, while RRT* had failed to converge to an optimal solution 50% of the times.

Fig 3 is another 2-D environment which is highly cluttered and PIB-RRT* was the quickest to find the optimal solution with the least number of average iterations to optimal solution ($i_{\text{avg}} = 20,539$) and it had the highest average rewiring per iteration rate ($\vartheta_{\text{PIB-RRT*}} = 1.25$). While PB-RRT* was second with ($i_{\text{avg}} = 22,124$) and ($\vartheta_{\text{PB-RRT*}} = 1.18$). Then came IB-RRT* and P-RRT* respectively in quickness too find the optimal solution while RRT* failed 50% of the times to find the optimal solution in this environment. Fig 4 shows a 3-D environment which is highly cluttered, having many obstacles between initial and goal regions. PIB-RRT* due to its $BPG()$ heuristic coupled with *Intelligent Sample Insertion* was able to find the optimal path the quickest ($i_{\text{avg}} = 17,430$). Then came PB-RRT* ($i_{\text{avg}} = 21,459$) and then IB-RRT* ($i_{\text{avg}} = 38,213$) and P-RRT* ($i_{\text{avg}} = 77,641$) while RRT* failed to find the optimal path. Fig 5 shows a 3-D environment with narrow passages through vertical columns and as seen from Table 1, PB-RRT* was the quickest in finding the optimal solution in this environment ($i_{\text{avg}} = 19,647$) with rewiring per iteration of ($\vartheta_{\text{PB-RRT*}} = 1.05$). PIB-RRT* was the second ($i_{\text{avg}} = 24,629$) while IB-RRT* ($i_{\text{avg}} = 39,583$) and P-RRT* ($i_{\text{avg}} = 83,523$) came third and fourth respectively in quickness to optimal solution and RRT* took an extraordinary number of iterations to converge to the optimal solution ($i_{\text{avg}} = 900,132$).

Fig 6 is a 3-D environment and has a highly obstacle ridden cluttered environment from initial to goal regions. We found PIB-RRT* to be the fastest to optimal solution in this environment ($i_{\text{avg}} = 131,542$) with highest rewiring per iteration rate ($\vartheta_{\text{PIB-RRT*}} = 1.31$) and PB-RRT* was second in quickness to optimal solution ($i_{\text{avg}} = 153,358$) with rewiring per iteration of ($\vartheta_{\text{PB-RRT*}} = 1.12$) and IB-RRT* ($i_{\text{avg}} = 223,493$), P-RRT* ($i_{\text{avg}} = 369,381$) and RRT* ($i_{\text{avg}} = 1,908,392$) came third, fourth and fifth respectively in quickness to optimal path. In this environment RRT* failed 60% of the times to reach an optimal solution. In Fig 7 different bar graphs are presented. These bar graphs have been obtained after experimentation and simulation in 10 different 2-D and 3-D environments for comparing PIB-RRT*, PB-RRT*, IB-RRT*, P-RRT* and RRT*. In Fig 7(a) the comparison is made in iterations required to find the initial feasible path in all environments for each Algorithm. Fig 7(b) shows the time required to find the initial feasible path for all the environments. Similarly Fig 7(c) and Fig 7(d) show the comparison of iterations required to find the optimal path and time required to find the optimal solution respectively. It is seen from Fig 7 that PIB-RRT* and PB-RRT* require very less iterations and time to find the initial path and the optimal solution when compared to IB-RRT*, P-RRT* and RRT*. It is noted that in some cases PIB-RRT* performs better than PB-RRT* and in some cases PB-RRT* performs better than PIB-RRT*. Mainly in highly cluttered environments PIB-RRT* has shown to perform better than PB-RRT* due to the inclusion of *IntelligentSampleInsertion* with $BPG()$ heuristic while in less cluttered environments PB-RRT* takes the lead in performance due to its greedy heuristic coupled with $BPG()$ heuristic. But this greedy heuristic becomes less efficient in highly cluttered environments.

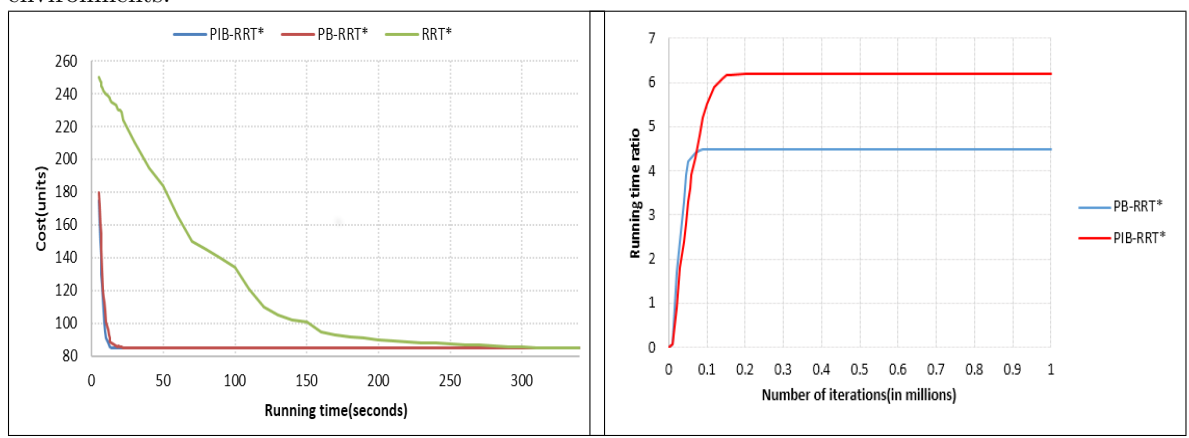

**Figure 8:** Cost vs Running Time         **Figure 9:** Running Time Ratio w.r.t RRT*

In Fig 8, after several runs in an obstacle filled 3D environment, the cost (in terms of Euclidean distance) versus the running time graph was plotted for PIB-RRT*, PB-RRT* and RRT* algorithms respectively. It can be seen that PIB-RRT* and PB-RRT* converge to optimal cost solution quite quickly as compared to RRT* due to $BPG()$ heuristic combined with bidirectional tree search. Fig 9 shows the running time ratio of PIB-RRT* over RRT* and PB-RRT* over

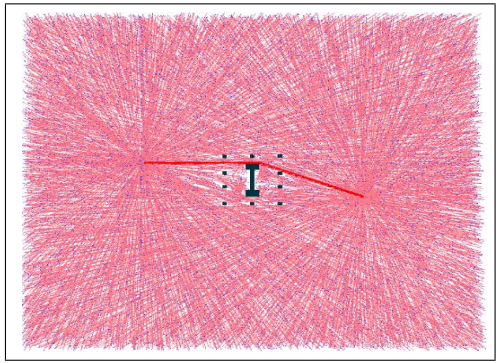
(a) PIB-RRT*: i=30000, k=1

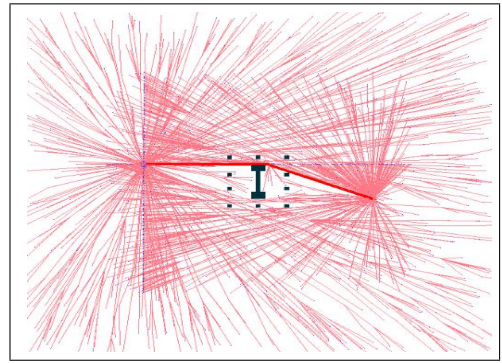
(b) PIB-RRT*: i=2500, k=300

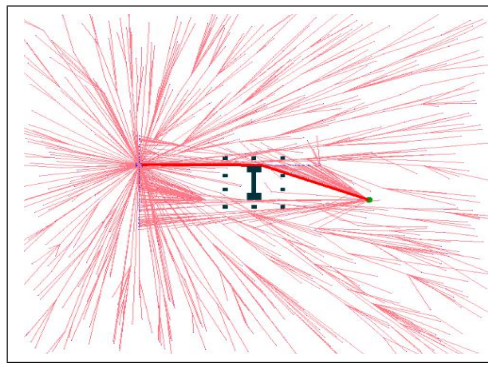
(c) PIB-RRT*: i=1500, k=600

**Figure 10:** Effect of $k$ on exploitation/exploration

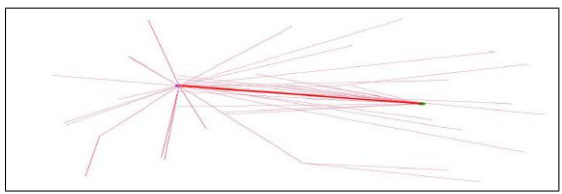
Figure 11: PIB-RRT*

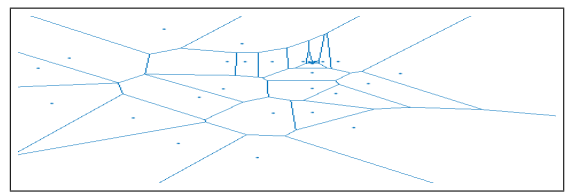
Figure 12: 30 Samples PIB-RRT*

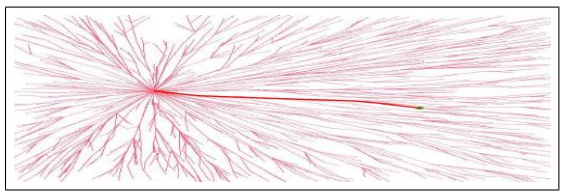
Figure 13: RRT*

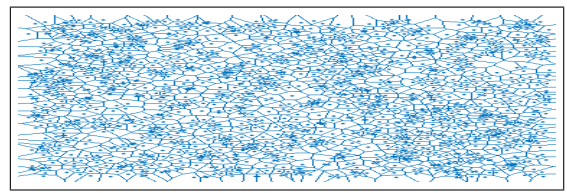
Figure 14: 1398 Samples RRT*

RRT* after numerous runs in an obstacle-free 3D environment. As expected from the computational complexity analysis of Section 5.4 and as stated in Theorem 6, Fig 9 shows that the running time ratios of PIB-RRT* and PB-RRT* w.r.t RRT* settle at a constant value as the number of iterations increase.

Fig 10 displays the effect of $k$ parameter of the $BPG()$ heuristic used in the proposed PIB-RRT* and PB-RRT* algorithms. The effects of $k$ parameter on exploitation and exploration can be clearly seen in Fig 10. As seen in the figure, lower values of $k$ biases our Algorithm towards more exploration while higher values of $k$ causes more exploitation. A balance has to be kept between exploration and exploitation for the proposed algorithms to work in all kinds of environments.

In Figure 11,12,13 and 14, PIB-RRT* and RRT* were run in the same environment till initial path was found. As seen in Fig 13 and Fig 14, RRT* took 1398 samples to find the initial solution path. The Voronoi biasing of RRT* is depicted in Fig 14, as observed it is quite uniform as no sample-biasing is present hence it took RRT* 1398 samples just to find the initial solution path and the resultant path had a non optimal cost. Where as in Fig 11, PIB-RRT* took only 30 samples to find the solution path in the same environment. Fig 12 shows the effect of $BPG()$ heuristic of PIB-RRT* on the Voronoi biasing. As seen the $BPG()$ heuristic potentially guides the random samples towards regions of higher intensity $I_z$ regions where rewiring rate is higher as stated in Theorem 4 and Theorem 5, hence the initial solution path was found in only 30 iterations and as seen in Fig 11, the initial path has optimal cost.

## 7. Conclusions and Future work

In this paper we have presented Bi-directional potential functions based asymptotically optimal, sampling path planning algorithms, PIB-RRT* and PB-RRT* which use $BPG()$ heuristic to potentially guide two Rapidly-exploring Random Trees towards each other, arguably the first of its kind, these algorithms PIB-RRT* and PB-RRT* have proven to be both theoretically and experimentally (**1**) similar in computational complexity as IB-RRT*, B-RRT* and RRT*; (**2**) provide asymptotic optimality; (**3**) avoids getting stuck in local minima as $APF$ does; (**4**) converges to optimal solution faster than its state of the art counter parts such as IB-RRT*, P-RRT* and RRT*; (**5**) lesser memory is consumed by PIB-RRT* and PB-RRT* as lesser iterations and time is required by them. By employing Bi-directional potential fields for the first time through fusing APF [19] into bi-directional variants of RRT* by using the proposed $BPG()$ heuristic, we have shown our proposed algorithms converge to optimal solution in the least amount of time and hence are of great importance in the physical real-time applications of motion planning of robots and online motion panniing of virtual characters and even can be used in nano robotics for surgery in the future.

In our future research, we plan to extend our algorithm for motion planning in dynamic environments [36] due to its rapid convergence to optimal solutions. Moreover, we also plan to leverage machine learning to cache feasible motion paths for experience-based motion planning that will enable our method to perform informed search for planning in new unseen environments.

## 8. Ackowledgements

The authors are greateful to Dr. Sertac Karaman of MIT for sharing the implementation of RRT* Algorithm.